\pdfoutput=1
\documentclass{article}

\usepackage{microtype}
\usepackage{graphicx}
\usepackage{subfigure}
\usepackage{booktabs} 

\usepackage[hyperindex,breaklinks]{hyperref}

\usepackage{algorithm}

\usepackage[accepted]{icml2020}

\icmltitlerunning{Learning with Bounded Instance- and Label-dependent Label Noise}

\usepackage{amsmath}
\usepackage[inline]{enumitem}

\usepackage{amsthm,amssymb,multirow,epstopdf,url,bm,dsfont}
\usepackage[skip=2pt]{caption}

\newtheorem{thm}{Theorem}
\newtheorem{defi}{Definition}
\newtheorem{lemm}{Lemma}
\newtheorem{cor}{Corollary}
\newtheorem{prop}{Proposition}
\newcommand{\tabincell}[2]{\begin{tabular}{@{}#1@{}}#2\end{tabular}}
\newcommand{\supp}{\mathrm{supp}}
\DeclareMathOperator{\sgn}{sgn}

\begin{document}

\twocolumn[
\icmltitle{Learning with Bounded Instance- and Label-dependent Label Noise}




\begin{icmlauthorlist}
\icmlauthor{Jiacheng Cheng}{usyd,ustc}
\icmlauthor{Tongliang Liu}{usyd}
\icmlauthor{Kotagiri Ramamohanarao}{unimelb}
\icmlauthor{Dacheng Tao}{usyd}
\end{icmlauthorlist}

\icmlaffiliation{usyd}{UBTECH Sydney AI Centre, School of Computer Science, Faculty of Engineering, University of Sydney, NSW, Australia.}
\icmlaffiliation{ustc}{University of Science and Technology of China, Hefei, China.}
\icmlaffiliation{unimelb}{School of Computing and Information Systems, University of Melbourne, VIC, Australia}

\icmlcorrespondingauthor{Tongliang Liu}{tongliang.liu@sydney.edu.au}

\icmlkeywords{Machine Learning, ICML}

\vskip 0.3in
]



\printAffiliationsAndNotice{This work was done when Jiacheng Cheng was a visiting student at University of Sydney.}

\begin{abstract}
Instance- and Label-dependent label Noise (ILN) widely exists in real-world datasets but has been rarely studied. In this paper, we focus on Bounded Instance- and Label-dependent label Noise (BILN), a particular case of ILN where the label noise rates---the probabilities that the true labels of examples flip into the corrupted ones---have upper bound less than $1$. Specifically, we introduce the concept of distilled examples, i.e. examples whose labels are identical with the labels assigned for them by the Bayes optimal classifier, and prove that under certain conditions classifiers learnt on distilled examples will converge to the Bayes optimal classifier. Inspired by the idea of learning with distilled examples, we then propose a learning algorithm with theoretical guarantees for its robustness to BILN. At last, empirical evaluations on both synthetic and real-world datasets show effectiveness of our algorithm in learning with BILN.
\end{abstract}

\section{Introduction}\label{sec1}


In the traditional classification task, we always expect and assume a perfectly labeled training sample. However, there is a strong possibility that we will be confronted with label noise, which means labels in the training sample are likely to be erroneous, especially in the era of big data. The reasons are as follows. On the one hand, to circumvent costly human labeling, many inexpensive approaches are employed to collect labeled data, such as data mining on social media and search engines \citep{fergus2010learning,schroff2011harvesting}, which inevitably involve label noise; on the other hand, even labels made by human experts are likely to be noisy due to confused patterns and perceptual errors. Overall, label noise is ubiquitous in real-world datasets and will undermine the performance of many machine learning models \citep{long2010random,frenay2014classification}. Therefore, designing learning algorithms robust to label noise is of significant value to the machine learning community \citep{SIGUA,yang2019searching,liu2019peer,xia2020parts,wu2020class2simi}.

There are several methods proposed to model label noise. The \emph{random classification noise}  (RCN) model, in which each label is flipped independently with a constant probability $\rho$, and the \emph{class-conditional random label noise} (CCN) model, in which the flip probabilities (noise rates) $\rho_{y}$ are the same for all labels from one certain class $y$, have been widely-studied \citep{angluin1988learning,kearns1998efficient,long2010random,gao2016on,han2018masking}.
A more generalized model is the \emph{instance- and label-dependent noise} (ILN), in which the flip rate $\rho_{y}(\mathbf{x})$ is dependent on both the instance $\mathbf{x}$ and the corresponding true label $y$. Obviously, the ILN model is more realistic and applicable. For example, in real-world datasets, an instance whose feature contains less information or is of poorer quality may be more prone to be labeled wrongly. Unfortunately, the case of ILN has not yet been extensively studied.

In this paper, label noise is defined to be \emph{Bounded Instance- and Label- dependent Noise} (BILN) if the noise rates for instances are upper bounded by some values smaller than $1$, and we focus on this situation. We propose an algorithm for learning with BILN and theoretically establish the statistical consistency and a performance bound. To the best of our knowledge, we are not aware of any other specially designed algorithms robust to such general label noise with theoretical guarantees. Empirical evaluations on synthetic and real-world datasets demonstrate the effectiveness of the proposed learning algorithm. In addition, we believe the proposed algorithm is promising to handle instance-dependent complementary label learning \citep{ishida2017learning,yu2018learning,xu2019generative, feng2019learning,chou2020unbiased}.

\textbf{Related Works: }
Learning with label noise has been widely investigated \citep{frenay2014classification}. There are lots of methods designed for learning with label noise. Some methods attempt to identify mislabeled training examples and then filter them out \citep{brodley1999identifying,zhu2003eliminating,angelova2005pruning,malach2017decoupling,jiang2018mentornet,han2018co,Kim_2019_ICCV,Huang_2019_ICCV,li2020dividemix}; some methods aim to modify existing learning models, e.g., deep learning models, without filtering the mislabeled examples out \citep{bylander1994learning,jin2003a,khardon2007noise,bootkrajang2012label-noise,bootkrajang2013boosting,goldberger2016training,ma2018dimensionality,ren2018learning,tanaka2018joint,hendrycks2018using,wang2018iterative,zhang2018generalized,xu2019l_dmi,li2019learning,yi2019probabilistic,Wang_2019_ICCV,nguyen2019self,hu2020simple}; some methods treat the unobservable true labels of training examples as hidden variables and learn them by maximum likelihood estimation \citep{lawrence2001estimating,bootkrajang2012label-noise,vahdat2017toward}. Most of these methods are heuristic and are not provided with theoretical guarantee for their robustness to label noise.

Many methods theoretically robust to RCN or CCN have been put forward with theoretical guarantees: \citet{natarajan2013learning} proposed two methods (the method of unbiased estimator and the method of label-dependent costs) to modify the surrogate loss and provided theoretical guarantees for the robustness to CCN of the modified loss; \citet{ghosh2015making} proved a sufficient condition for a loss function to be robust to symmetric CCN; \citet{ghosh2017robust} extended the results in \citet{ghosh2015making} to multiclass classification. \citet{van2015learning} proved that the unhinged loss is the only convex loss that is robust to symmetric CCN;  \citet{patrini2016loss} introduced linear-odd losses and proved that every linear-odd loss is approximately robust to CCN; \citet{liu2016classification} proved that by importance reweighting, any loss function can be robust to CCN.
\citet{northcutt2017rankpruning} proposed the method of rank pruning to estimate noise rates and remove mislabeled examples prior to training. 
Many methods \citep{natarajan2013learning,yu2017transfer} employ noise rates (transition probabilities) to produce noise-robust loss functions. \citet{liu2016classification,menon15,ramaswamy2016mixture} and \citet{scott2015rate} provided consistent estimators for the noise rates 
and the inversed noise rates, respectively.
\citet{patrini2017making} extended the estimators of \citet{liu2016classification,menon15} to the multi-class setting. \citet{xia2019anchor} proposed a noise rate estimator for which anchor points are no longer necessary. Recently, \citet{liu2019peer} presented peer loss functions that operate with noisy labels without the need of specifying the class-conditional noise rates.


Especially, many advances have been achieved in learning halfspaces in the presence of different degrees of label noise \citep{Awasthi15b,awasthi16,zhang17b,Yan17,diakonikolas2019distribution}. These works often assume examples of two classes to be linearly separable under the clean distribution and the marginal over $\mathbf{X}$ to have some special structures (e.g.\ uniform over the unit sphere).

Learning with more realistic label noise has also been studied in recent years \citep{xiao2015learning,li2017learning,lee2017cleannet,tanaka2018joint,seo2019combinatorial,Han_2019_ICCV}.
These works have been evaluated on real-world label noise, but theoretical guarantees for noise-robustness have not been provided.

For ILN, \citet{menon2018learning} proved that, in the special case where $\rho_{+1}(\mathbf{x})=\rho_{-1}(\mathbf{x})$, the Bayes optimal classifiers under the clean and noisy distributions coincide, implying that any algorithm consistent for the classification under the noisy distribution is also consistent for the classification under the clean distribution. \citet{xia2020parts} tried to address the instance-dependent label noise by exploiting the the parts-dependence assumption. To the best of our knowledge, we are not aware of any algorithm prior to this work dealing with ILN with theoretical guarantees.

\textbf{Organization: }
The rest of this paper is structured as follows. In Sec.\ \ref{sec3}, we formalize our research problem. In Sec.\ \ref{sec5}, our algorithm for learning with BILN is presented in detail. In Sec.\ \ref{sec6}, we provide empirical evaluations of our learning algorithm. In Sec.\ \ref{sec7}, we conclude our paper. All the proofs are presented in the supplementary material.

\section{Problem Setup}\label{sec3}
In the task of binary classification with label noise, we consider a feature space $\mathcal{X} \subset \mathbb{R}^{d}$ and a label space $\mathcal{Y} = \{-1, +1\}$. Formally, we assume that random variables $(\mathbf{X},Y,\widetilde{Y}) \in \mathcal{X} \times \mathcal{Y} \times \mathcal{Y}$ are jointly distributed according to an unknown distribution $P$, where $\mathbf{X}$ is the observation, $Y$ is the uncorrupted but unobserved label and $\widetilde{Y}$ is the observed but noisy label. In specific, we use $D$ and $D_{\rho}$ to denote the clean distribution $P(\mathbf{X},Y)$ and the noisy distribution $P(\mathbf{X},\widetilde{Y})$, respectively.
With label noise, we observe a sequence of pairs $\{(\mathbf{x}_{i},\widetilde{y}_{i})\}_{i=1}^n$ sampled i.i.d.\ from $D_{\rho}$ and our goal is to construct a discriminant function $f$ : $\mathcal{X}\rightarrow\mathbb{R}$, such that the classifier $g(\mathbf{x})=\sgn(f(\mathbf{x}))$ is an accurate prediction of the label of $\mathbf{x}$, where $\sgn(\cdot)$ denotes the sign function. Some criteria are necessary to measure the performance of $f$ and $g$. In the first place, we have the \emph{0-1 risk} of $g$ as
\[
R_{D}(g) =\mathbb{E}_{(\mathbf{X},Y)\sim D} [\mathds{1}[g(\mathbf{X}) \neq Y]]
\]
where $\mathbb{E}$ denotes expectation and its subscript indicates the random variables and the distribution w.r.t.\ which the expectation is taken, and $\mathds{1}[\cdot]$ denotes the indicator function. Then we can define the Bayes optimal classifier under $D$ as
$
g_{D}^{*}=\mathop{\arg\min}_{g}R_{D}(g)
$
and the Bayes risk $R_{D}^* = \mathop{\min}_{g}R_{D}(g)$.
Since the distribution $D$ is unknown to us, we cannot directly compute $R_{D}(g)$. So we need the \emph{empirical 0-1 risk} and a sample $\{(\mathbf{x}_{i},{y}_{i})\}_{i=1}^n$ sampled i.i.d.\ from $D$ to estimate $R_{D}(g)$:
\[
\widehat{R}_{D}(g) = \frac{1}{n} \sum^{n}_{i=1} \mathds{1}[g(\mathbf{x}_{i})\neq y_{i}].
\] NP-hardness of minimizing the 0-1 risk, which is neither convex nor smooth, forces us to adopt surrogate loss functions \citep{bartlett2006convexity,scott2012calibrated}. When the surrogate loss function $L(f(\mathbf{X}),Y)$ is used, we define the \emph{$L$-risk} of $f$ as
\[
R_{D,L}(f)=\mathbb{E}_{(\mathbf{X},Y)\sim D}[L(f(\mathbf{X}),Y)].
\]
If $L$ is classification-calibrated,
the minimizer of $L$-risk (if it exists) $f^{*}_{D,L}=\mathop{\arg\min}_{f}R_{D,L}(f)$ will also minimizes the 0-1 risk, i.e., $ R_{D}(\sgn(f^{*}_{D,L})) = R_{D}^{*}$ \citep{bartlett2006convexity}. Likewise, the \emph{empirical L-risk} is defined to estimate the $L$-risk:
\[ \widehat{R}_{D,L}(f) = \frac{1}{n} \sum^{n}_{i=1} L(f(\mathbf{x}_{i}),y_{i}). \]
Risks under the noisy distribution can be defined similarly as risks under the clean distribution.

As for label noise, we employ the noise rate $\rho_{y}(\mathbf{x})\!=\!P(\widetilde{Y}\!=\!-y|\mathbf{X}\!=\!\mathbf{x},Y\!=\!y)$ to model it. The noise is said to be random classification noise (RCN) if $\rho_{+1}(\mathbf{x})\!=\!\rho_{-1}(\mathbf{x})\!=\!\rho$ or class-conditional random label noise (CCN) if $\rho_{y}(\mathbf{x})$ is independent on $\mathbf{x}$ but dependent on $y$. A more general model of label noise is instance- and label-dependent noise (ILN). For ILN, $\rho_{y}(\mathbf{x})$ is dependent on both the observation $X$ and the true label $Y$. The model of ILN is more realistic and applicable because, e.g., observations with misleading contents are more likely to be annotated with wrong labels.

This paper focus on a particular case of ILN where noise rates have upper bounds. Formally, the noise is said to be bounded instance- and label- dependent noise (BILN) if the following assumption holds.
\newtheorem{Assumption}{Assumption}
\begin{Assumption}\label{as1}
$\forall \mathbf{x}\in\mathcal{X}$, we have
\[\left\{
\begin{aligned}
0\leq&\rho_{+1}(\mathbf{x})\leq\rho_{\text{{+1max}}}<1,\\
0\leq&\rho_{-1}(\mathbf{x})\leq\rho_{\text{{-1max}}}<1,\\
0\leq&\rho_{+1}(\mathbf{x})+\rho_{-1}(\mathbf{x})<1.\\
\end{aligned}
\right.
\]

\end{Assumption}
The bounded noise rate assumption $0\leq\rho_{+1}(\mathbf{x})+\rho_{-1}(\mathbf{x})<1$ encodes that for each example, the noisy label and clean label must agree on average \citep{menon2018learning}. Especially when noise rates are only dependent on labels, $\rho_{\text{+1}}+\rho_{\text{-1}}<1$ is a standard condition for analysis under CCN \citep{blum1998combining, natarajan2013learning}. In the rest of this paper, we always suppose Assumption \ref{as1} holds.

Notice that our BILN model is different from the bounded noise model (a.k.a.\ the Massart noise model \citep{massart2006risk}), which assumes that $P_{D}(\mathbf{X}=\mathbf{x}|Y\!=\!+1)$ and $P_{D}(\mathbf{X}=\mathbf{x}|Y\!=\!-1)$ have non-overlapping supports and that noise rates are upper bounded by a constant smaller than 0.5.

\section{Learning with BILN}\label{sec5}
In this section, we propose an algorithm, inspired by the idea of learning with \emph{distilled examples} (explained in the following subsection, i.e., examples whose labels are identical with the labels assigned for them by $g_{D}^{*}$, the Bayes optimal classifier under the clean distribution), for learning with bounded instance- and label-dependent label noise. Recently, a similar idea has been applied for learning with class-conditional noise \citep{zheng2020Error}.

This section is structured as follows. In Sec.\ \ref{sub1}, we prove that under certain conditions classifier learnt on distilled examples converge to $g^{*}_{D}$. In Sec.\ \ref{sub2}, an automatic approach is proposed to collect distilled examples out of noisy examples utilizing the knowledge of $\rho_{\text{+1max}}$ and $\rho_{\text{-1max}}$. In Sec.\ \ref{sub3}, we discuss the necessity of actively labeling a small fraction of noisy examples for our learning algorithm to be statistically consistent. In Sec.\ \ref{sub4}, we further employ importance reweighting to prevent our learning algorithm from suffering from sample selection bias. In Sec.\ \ref{sub553}, an approach to collect distilled examples without knowledge of $\rho_{ \text{{+1max}}}$ and $\rho_{\text{{-1max}}}$ is proposed.

Note that, to simplify analysis, in this section we assume  upper bounds of noise rates $\rho_{ \text{{+1max}}},\rho_{\text{{-1max}}}$ to be known to us until Sec.\ \ref{sub553}. We propose an approach to collect distilled examples without knowledge of $\rho_{ \text{+1max}},\rho_{\text{-1max}}$ in Sec.\ \ref{sub553} which can be easily integrated into our algorithm, and empirical results in Sec.\ \ref{sec72} and Sec.\ \ref{sec73} demonstrate that our algorithm can work well with or without knowing $\rho_{\text{{+1max}}}$ and $\rho_{\text{{-1max}}}$.

\subsection{Learning with Distilled Examples}\label{sub1}
We formally introduce the concept of distilled example first.
\begin{defi}
An example $(\mathbf{x},y)$ is defined to be a distilled example if its label is identical to the one assigned by the Bayes optimal classifier under the clean data, i.e., $y=g_D^{*}(\mathbf{x})$.
\end{defi}

Denote by $D^{*}$ the distribution of distilled examples. In the empirical risk minimization (ERM) frame, a discriminant function $\hat{f}_{D^{*},L}$ can be learnt by
\[
\hat{f}_{D^{*},L}=\mathop{\arg\min}_{f\in\mathcal{F}}\widehat{R}_{D^{*},L}(f),
\]
where $\mathcal{F}$ is the learnable function class. We will show that under certain conditions, $\sgn(\hat{f}_{D^{*},L})$ converges to $g_D^{*}$, the Bayes optimal classifier under the clean distribution. Before presenting the main theoretical results, we introduce the following lemma and theorem.


\begin{lemm}\label{lm2}
Denote by $\eta(\mathbf{x})$  the conditional probability $P_{D}(Y\!=\!+1|\mathbf{X}=\mathbf{x})$. The Bayes optimal classifier under $D$ is given by $g^{*}_{D}(\mathbf{x})=\sgn\left(\eta(\mathbf{x})-\frac{1}{2}\right)$.
\end{lemm}

\begin{thm}\label{th2}
Given the target distribution $D$ and the distilled examples' distribution $D^{*}$. If marginal distributions $P_{D}(\mathbf{x})$ and $P_{D^{*}}(\mathbf{x})$ share the same support, then the Bayes optimal classifier under $D^{*}$ coincides with the Bayes optimal classifier under $D$, i.e.\ $g^{*}_{D^{*}}= g^{*}_{D}$.
\end{thm}

Combining the aforementioned results with the basic Rademacher bound \citep{bartlett2002rademacher}, we have the following proposition.
\begin{prop}
Under the condition of Theorem \ref{th2}, assume that $\{(\mathbf{x}_{i},y_{i})\}^{m}_{i=1}$ are i.i.d.\ sampled from $D^*$. If $L$ is $[0,b]$-valued and $f^{*}_{D^*,L}=\mathop{\arg\min}_{f}R_{D^*,L}(f)\in\mathcal{F}$, then for any $\delta$, with probability at least $1-\delta$, we have
\begin{align*}
R_{D^{*},L}(\hat{f}_{D^{*},L})-R_{D^{*},L}(f^{*}_{D^*,L})\\
\leq2\mathfrak{R}(L\circ\mathcal{F})+2b\sqrt{\frac{log(1/\delta)}{2m}},
\end{align*}
where the Rademacher complexity
$\mathfrak{R}(L\circ\mathcal{F})\!=\!\mathbb{E}_{D^*,\sigma}[\mathop{\sup}_{f\in\mathcal{F}}\frac{2}{m}\sum \limits_{i=1}^{m}\sigma_{i}L(f(\mathbf{x}_{i},y_{i}))].$
($\sigma_{1},\cdots,\sigma_{m}$ are independent Rademacher variables.)
\end{prop}
The above proposition implies that $R_{D^{*},L}(\hat{f}_{D^{*},L})$  converges to $R_{D^{*},L}(f^{*}_{D^*,L})$, as $m\to\infty$. Further, if $L$ is classification-calibrated and the Bayes optimal classifier is within the predefined $\mathcal{F}$, $\sgn(\hat{f}_{D^*,L})$ will converge to the Bayes optimal classifier under $D^*$, which is also the Bayes optimal classifier under $D$, as the number of distilled examples approaches infinity.

Motivated by above results, we discuss how to collect distilled examples out of noisy examples and learn a well-performing classifier with distilled examples in the following subsections.

\subsection{Collecting Distilled Examples out of Noisy Examples Automatically}\label{sub2}

Then we propose an approach to automatically collect distilled examples out of noisy examples according to the following theorem and its immediate corollary.
\begin{thm}\label{th5}
Denote by $\tilde{\eta}(\mathbf{x})$  the conditional probability $P_{D_\rho}(\widetilde{Y}\!=\!+1|\mathbf{X}\!=\!\mathbf{x})$.
$\forall \mathbf{x}\in\mathcal{X}$, given that $UB(\rho_{\pm1}(\mathbf{x}))$ is an upper bound of $\rho_{\pm1}(\mathbf{x})$, we have\\
$\tilde{\eta}(\mathbf{x})<\frac{1-UB(\rho_{+1}(\mathbf{x}))}{2}\implies(\mathbf{x},Y\!=\!-1)$ is distilled; \\ $\tilde{\eta}(\mathbf{x})>\frac{1+UB(\rho_{-1}(\mathbf{x}))}{2}\implies(\mathbf{x},Y\!=\!+1)$ is distilled.
\end{thm}

\begin{cor}\label{th3}
$\forall \mathbf{x}\in\mathcal{X}$, we have \\ $\tilde{\eta}(\mathbf{x})<\frac{1-\rho_{\text{+1max}}}{2}\implies(\mathbf{x},Y\!=\!-1)$ is distilled; \\ $\tilde{\eta}(\mathbf{x})>\frac{1+\rho_{\text{-1max}}}{2}\implies(\mathbf{x},Y\!=\!+1)$ is distilled.
\end{cor}

According to Corollary \ref{th3}, we can obtain distilled examples by picking out every noisy example $(\mathbf{x}_{i},\widetilde{y}_{i})$ whose $\mathbf{x}_{i}$ satisfies $\tilde{\eta}(\mathbf{x}_{i})>\frac{1+\rho_{\text{-1max}}}{2}$ or $\tilde{\eta}(\mathbf{x}_{i})<\frac{1-\rho_{\text{+1max}}}{2}$ and then  assigning the label $+1$ or $-1$ to it.

Indeed, in practice $\tilde{\eta}$ is inaccessible to us, but it is feasible for us to obtain an estimator $\hat{\tilde{\eta}}$ for $\tilde{\eta}$. Note that the estimation of $\tilde{\eta}$ is a traditional probability estimation problem which can be addressed by several methods, such as the probabilistic classification methods (e.g.,\ logistic regression, deep neural networks), the kernel density estimation methods, and the density ratio estimation methods.

\subsection{Labeling Noisy Examples Actively}\label{sub3}
The collection of distilled examples in the last subsection is inevitably biased, because examples whose observations are in $\left\{\mathbf{x}\in\mathcal{X}|\frac{1-\rho_{\text{{+1max}}}}{2}\leq\tilde{\eta}(\mathbf{x})\leq\frac{1+\rho_{\text{{-1max}}}}{2}\right\}$ will not be collected. To put it more formally, we let $D^{*}_{\text{auto}}$ denote the distribution of these  distilled examples automatically collected via Corollary \ref{th3}. Then we have 
\begin{align*}
&\supp\left(P_{D^{*}_{\text{auto}}}(\mathbf{x})\right)\\
=&\left\{\mathbf{x}\!\in\!\mathcal{X}|\tilde{\eta}(\mathbf{x})\in\left[0,\frac{1-\rho_{\text{+1max}}}{2}\right)\bigcup
\left(\frac{1+\rho_{\text{-1max}}}{2},1\right]\right\}
\end{align*}
which leads to $\supp(P_{D^{*}_{\text{auto}}}(\mathbf{x}))\neq\supp(P_{D}({\mathbf{x}}))$ and does not hold the condition of Theorem \ref{th2}. Consequently, learning with automatically-collected distilled examples only is not statistically consistent.

Our strategy to address this issue is to perform active learning. Formally, learning algorithms which actively choose unlabeled examples, manually acquire their labels and then use labeled examples to perform supervised learning are called active learning methods \citep{settles2010active}. Active learning has been successfully applied in many fields, e.g., text classification \citep{tong2001support2} and compound classification \citep{lang2016feasibility}. Also, it is applicable to our case, where we treat the automatically-collected distilled examples as labeled data and the remaining noisy examples as unlabeled data since their labels are noisy and unreliable. We will have some of the remaining noisy examples labeled by human experts and train the classifier using the automatically-collected and actively-labeled distilled examples together. Here we made a reasonable assumption that actively-labeled examples are distilled as well, i.e., manual labels by human experts are the same with labels given by the Bayes optimal classifier under clean distribution. 

As for how to determine which examples to be actively labeled, we adopt a simple but widely-used strategy: Choose unlabeled examples at random, which ensures that $\supp\left(P_{D^{*}}(\mathbf{x})\right)=\supp\left(P_{D}(\mathbf{x})\right)$, and further makes our learning algorithm  consistent.

\subsection{Covariate Shift Correction by Importance Reweighting}\label{sub4}
In previous subsections, we introduced our approach to construct a sample of distilled examples $\{(\mathbf{x}_{i}^{\text{distilled}},y_{i}^{\text{distilled}})\}^{m}_{i=1}$ and show that learning on the distilled sample is consistent. However, in practice, we noticed that the performance of the classifier directly learnt on the distilled examples is likely to be comprised by the problem of sample selection bias, because the distribution of distilled examples $D^{*}$ does not exactly match the target distribution $D$. 

Technically, the sample selection bias can be divided into the difference between $P_{D}(\mathbf{x})$ and $P_{D^{*}}(\mathbf{x})$ and the difference between $P_{D}(y|\mathbf{\mathbf{x}})$ and $P_{D^{*}}(y|\mathbf{x})$. Here we focus on the former for the following reasons. First, according to our analysis in Sec.\ \ref{sub1}, the bias in $P(y|\mathbf{x})$ in our case does not change the Bayes optimal classifier. Second, the bias in $P(\mathbf{x})$ in our case is severe because the number of actively-labeled examples is usually extremely limited since manual labeling is costly. Consequently, the proportion of examples whose observations are in $\{\mathbf{x}\in\mathcal{X}|\tilde{\eta}(\mathbf{x})\in[\frac{1-\rho_{\text{{+1max}}}}{2},
\frac{1+\rho_{\text{{-1max}}}}{2}]\}$ in our distilled sample is significantly smaller compared to that in a sample from $D$. Hence, we make the following assumption.
\begin{Assumption}\label{as2}
$P_{D}(\mathbf{x},y)$ and $P_{D^{*}}(\mathbf{x},y)$ only differ in the marginal distribution  $P(\mathbf{x})$.
\end{Assumption}
Then the issue of sample selection bias is simplified as \emph{covariate shift}. Importance reweighting is a method to handle this problem as follows.

\begin{align}
R_{D,L}(f)&=\mathbb{E}_{(\mathbf{X},Y)\sim D}[L(f(\mathbf{X}),Y)]\notag\\
&=\mathbb{E}_{(\mathbf{X},Y)\sim D^{*}}[\frac{P_{D}(\mathbf{X},Y)}{P_{D^{*}}(\mathbf{X},Y)}L(f(\mathbf{X}),Y)]\notag\\
&=\mathbb{E}_{(\mathbf{X},Y)\sim D^{*}}[\frac{P_{D}(\mathbf{X})}{P_{D^{*}}(\mathbf{X})}L(f(\mathbf{X}),Y)]\notag\\
&=\mathbb{E}_{(\mathbf{X},Y)\sim D^{*}}[\beta(\mathbf{X})L(f(\mathbf{X}),Y)]\ \notag\\
&=R_{D^{*},\beta L}(f),\label{eq4}
\end{align}
where the second equality follows by Assumption \ref{as2} and the importance $\beta(\mathbf{x})=\frac{P_{D}(\mathbf{x})}{P_{D^{*}}(\mathbf{x})}$.
Eq.\ (\ref{eq4}) implies that given $\beta$, we can minimize $R_{D,L}(f)$ by minimizing $R_{D^{*},\beta L}(f)$ which can be estimated as
\[
\widehat{R}_{D^{*},\beta L}(f)=\frac{1}{m}\sum^{m}_{i=1}\beta(\mathbf{x}^{\text{distilled}}_i)L(f(\mathbf{x}^{\text{distilled}}_i),y^{\text{distilled}}_i). \]
Further, $\hat{f}_{D^{*},\beta L}$ can be learnt by
\begin{equation}
\hat{f}_{D^{*},\beta L}=\mathop{\arg\min}\limits_{f\in\mathcal{F}}\widehat{R}_{D^{*},\beta L}(f).\label{ERM_reweight}
\end{equation}
A performance bound for $\hat{f}_{D^{*},\beta L}$ is provided in Proposition \ref{prop2}.
\begin{prop}\label{prop2}
Assume that Assumption \ref{as2} holds and $\{(\mathbf{x}_{i},y_{i})\}^{m}_{i=1}$ are i.i.d.\ sampled from $D^*$, which satisfies that $\supp{\left(P_{D*}(\mathbf{x})\right)}=\supp{\left(P_{D}(\mathbf{x})\right)}$. If $\beta(\mathbf{x})L(f(\mathbf{x}),y)$ is [0,b]-valued and $f^*_{D,L}=\mathop{\arg\min}_{f}R_{D,L}(f)\in\mathcal{F}$, then for any $\delta>0$, with probability at least $1-\delta$, we have
\[
R_{D,L}(\hat{f}_{D^{*},\beta L})-R_{D,L}(f^{*}_{D,L})
\leq
\]
\[
2\mathfrak{R}(\beta\circ L\circ\mathcal{F})+2b\sqrt{\frac{log(1/\delta)}{2m}}
,
\]
where the Rademacher complexity $\mathfrak{R}(\beta\circ L\circ\mathcal{F})=\mathbb{E}_{D^*,\sigma}[\mathop{\sup}_{f\in\mathcal{F}}\frac{2}{m}\sum\limits_{i=1}^{m}\sigma_{i}\beta(\mathbf{x}_{i}) L(f(\mathbf{x}_{i}),y_{i})].
$
\end{prop}
The method of empirical kernel mean matching (KMM) \citep{huang2007correcting} can be employed to estimate the importance. By KMM, given two sets of observations $\{\mathbf{x}_{i}\}^{n}_{i=1}$ and $\{\mathbf{x}^{\text{distilled}}_{i}\}^{m}_{i=1}$ sampled from $P_{D}(\mathbf{x})$ and $P_{D^{*}}(\mathbf{x})$ respectively, we can obtain proper importance $\bm{\beta}=[\beta_{1},\cdots,\beta_{m}]=\left[\beta(\mathbf{x}^\text{distilled}_{1}),\cdots,\beta(\mathbf{x}^{\text{distilled}}_{m})\right]$  via solving
\begin{equation}
\label{opt1}
\begin{aligned}
&\mathop{\text{minimize}}_{\beta} &&\frac{1}{2}\bm{\beta}^{T}\bm{K}\bm{\beta}-\bm{\kappa}^{T}\bm{\beta},\\
&\text{subject to} &&\text{$\forall i, \beta_{i}\in[0,B] $ and $|\sum_{i=1}^{m}\beta_{i}-m|\leq m\epsilon$,}
\end{aligned}
\end{equation}
where $\bm{K}_{ij}\!=\!k\left(\mathbf{x}^{\text{distilled}}_{i},\mathbf{x}^{\text{distilled}}_{j}\right)$, $\bm{\kappa}_{i}\!=\!\frac{m}{n}\sum_{j=1}^{n}k(\mathbf{x}^{\text{distilled}}_{i},\mathbf{x}_{j})$ and $k$ is a universal kernel. Optimization problem (\ref{opt1}) is a quadratic program which can be solved efficiently using interior point methods or any other successive optimization procedure.

Finally, our learning algorithm is summarized in Algorithm \ref{algo2}. The proposed framework can be easily extended to multiclass classification (cf.\ Supplementary Material B). 

Admittedly, in order for the analysis in this subsection to be simplified, the target distribution $D$ is required to satisfy Assumption \ref{as2}, which might not be valid in certain cases. In addition, the time complexity of the KMM procedure might be a potential concern. Remind that our learning algorithm is consistent, no matter whether Assumption \ref{as2} is satisfied and whether importance reweighting is performed. We present covariate shift correction by importance reweighting as a part of our algorithm mainly because we observed that it usually boosts the algorithm performance in practice.

\subsection{Collecting Distilled Examples Without Knowledge of $\rho_{\text{{+1max}}}$ and $\rho_{\text{{-1max}}}$}\label{sub553}
In the previous subsections we assumed $\rho_{+1\text{max}},\rho_{-1\text{max}}$ to be known to us, which is a strong assumption and seldom holds in real-world tasks.  In order to make our algorithm more practical, we propose an approach to collect distilled examples without knowledge of $\rho_{+1\text{max}}$ and $\rho_{-1\text{max}}$ in this section.

To collect distilled examples without knowledge of $\rho_{\pm1\text{max}}$ by Theorem \ref{th5}, we need to find $UB(\rho_{\pm1}(\mathbf{x}))$ using only noisy examples.
We have the following theorem that provides an upper bound for $\rho_{\pm1}(\mathbf{x})$.
\begin{thm}\label{th6}
$\forall \mathbf{x}\in\mathcal{X}$, we have $\rho_{\text{+1}}(\mathbf{x}) \leq 1-\tilde{\eta}(\mathbf{x})$ and $\rho_{\text{-1}}(\mathbf{x}) \leq \tilde{\eta}(\mathbf{x})$.
\end{thm}

\begin{algorithm}[t]
\caption{Learning with BILN}
\label{algo2}
\textbf{Input}: the noisy sample $S_{\text{noisy}}\!=\!\{(\mathbf{x}_{i},\tilde{y}_{i})\}^{n}_{i=1}$, the upper bounds of noise rates $\rho_{\text{{+1max}}}$, $\rho_{\text{{-1max}}}$, the number of examples to be actively labeled $n_\text{act}$;


\begin{algorithmic}[1]
\STATE Initialize the distilled sample $S_{\text{distilled}}=\emptyset$;
\STATE Learn $\hat{\tilde{\eta}}$ on $S_{\text{noisy}}$;
\FOR{$(\mathbf{x}_{i},\tilde{y}_{i})$ in $S_{\text{noisy}}$}
\IF{$\hat{\tilde{\eta}}(\mathbf{x}_{i})>\frac{1+\rho_{\text{{-1max}}}}{2}$}
\STATE $S_{\text{distilled}}\gets S_{\text{distilled}}\!\cup\!\{(\mathbf{x}_{i},y^{\text{distilled}}_i=+1)\}$; $S_{\text{noisy}}\gets S_{\text{noisy}}\!\setminus\!\{(\mathbf{x}_{i},\tilde{y}_{i})\}$;
\ENDIF
\IF{$\hat{\tilde{\eta}}(\mathbf{x}_{i})<\frac{1-\rho_{\text{{+1max}}}}{2}$}
\STATE $S_{\text{distilled}} \gets S_{\text{distilled}}\!\cup\!\{(\mathbf{x}_{i},y^{\text{distilled}}_i=-1)\}$; $S_{\text{noisy}}\gets S_{\text{noisy}}\!\setminus\!\{(\mathbf{x}_{i},\tilde{y}_{i})\}$;
\ENDIF
\ENDFOR
\STATE Randomly sample $n_\text{act}$ examples $\{(\mathbf{x}^{\text{act}}_{i},\tilde{y}^{\text{act}}_{i})\}_{i=1}^{n_\text{act}}$ from $S_\text{noisy}$;
\FOR{$i=1$ to $n_\text{act}$}
\STATE Query $\mathbf{x}^{\text{act}}_{i}$ for its distilled label $y^{\text{act}}_{i}$;
\STATE $S_{\text{distilled}}\gets S_{\text{distilled}}\!\cup\!\left\{(\mathbf{x}^{\text{act}}_{i},y^{\text{act}}_{i})\right\}$;
\ENDFOR
\STATE Obtain importance of distilled examples in $S_{\text{distilled}}$ via solving (\ref{opt1});
\STATE Learn $\hat{f}$ on the distilled sample $S_{\text{distilled}}$ by (\ref{ERM_reweight});
\end{algorithmic}
\end{algorithm}

Note that $1-\tilde{\eta}(\mathbf{x})$ and $\tilde{\eta}(\mathbf{x})$ cannot be directly used as $UB(\rho_{+1}(\mathbf{x}))$ and $UB(\rho_{-1}(\mathbf{x}))$ for collecting distilled examples by Theorem \ref{th5}, since $\tilde{\eta}(\mathbf{x})<\frac{\tilde{\eta}(\mathbf{x})}{2}$ and $\tilde{\eta}(\mathbf{x})>\frac{1+\tilde{\eta}(\mathbf{x})}{2}$ can never be satisfied. 

Our strategy is to consider the $k$-nearest neighborhood $\mathcal{N}_k(\mathbf{x})$ of a given example $\mathbf{x}$ in the feature space, and use $\sum_{\mathbf{x}_j\in \mathcal{N}_k(\mathbf{x})}\frac{\tilde{\eta}(\mathbf{x}_j)}{k}$ and $\sum_{\mathbf{x}_j\in \mathcal{N}_k(\mathbf{x})}\frac{1-\tilde{\eta}(\mathbf{x}_j)}{k}$, which can be estimated using $\hat{\tilde{\eta}}$, as approximate upper bounds of $\rho_{+1}(\mathbf{x})$ and $\rho_{-1}(\mathbf{x})$. Experimental results demonstrate that integrating this approach of collecting distilled examples into Algorithm \ref{algo2} can achieve decent results and it is robust to the selection of $k$.

\section{Empirical Evaluations}\label{sec6}
\begin{figure*}[!t]
\centering
\subfigure[]{
  \begin{minipage}[t]{0.235\linewidth} %
    \centering
    \includegraphics[width=\linewidth]{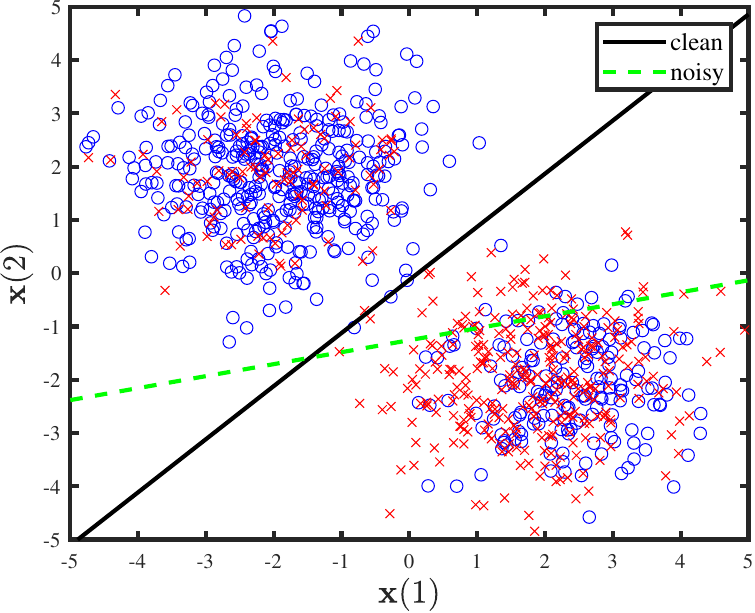}
    \label{vis1}
  \end{minipage}
  }
\subfigure[]{
  \begin{minipage}[t]{0.235\linewidth} %
    \centering
    \includegraphics[width=\linewidth]{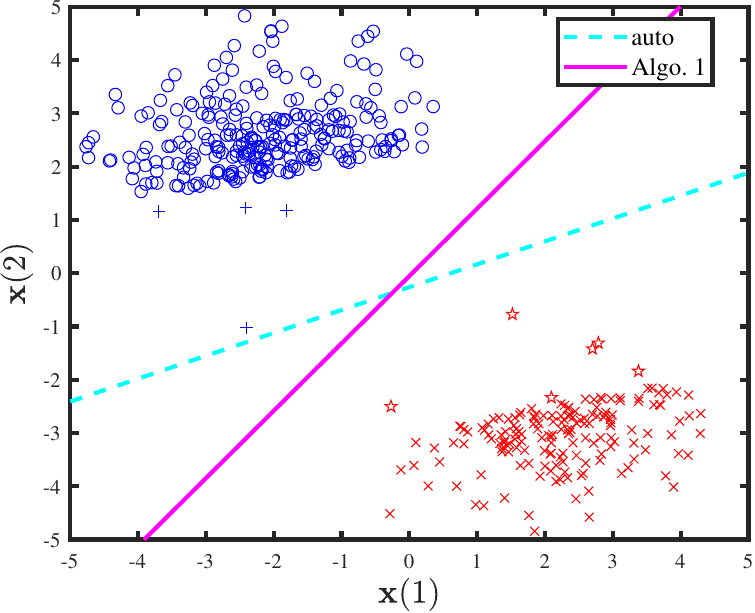}
    \label{vis2}
  \end{minipage}
  }
\subfigure[]{
  \begin{minipage}[t]{0.235\linewidth} %
    \centering
    \includegraphics[width=\linewidth]{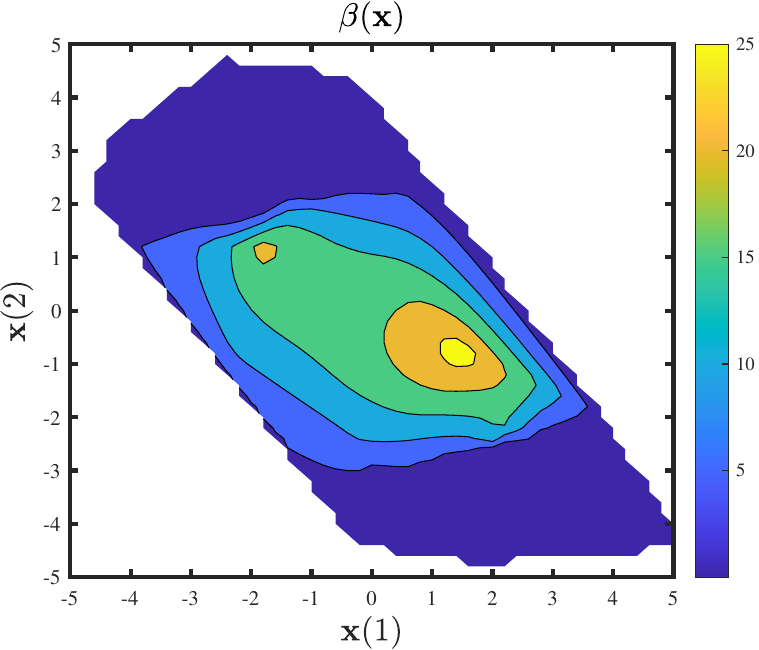}
    \label{vis3}
  \end{minipage}
  }
\subfigure[]{
  \begin{minipage}[t]{0.235\linewidth} %
    \centering
    \includegraphics[width=\linewidth]{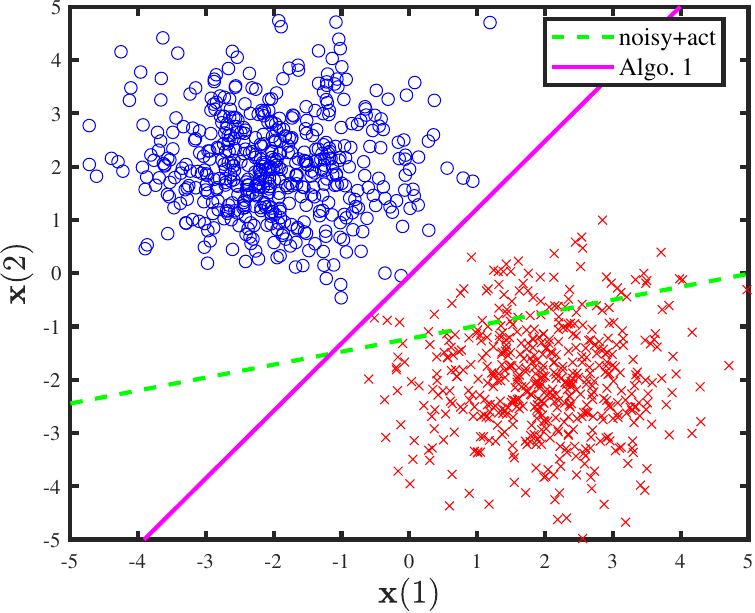}
    \label{vis4}
  \end{minipage}
  }
\caption{Visualization of the procedure of our Algorithm \ref{algo2} (best viewed in color). In this trial, ($\rho_{\text{+1max}}$, $\rho_{\text{-1max}}$, $n_\text{act}$)=(0.25, 0.49, 10) and $\mathbf{w}_{+1}=[-0.2723, -0.8796, 0.4133]^{\top}$, $\mathbf{w}_{-1}=[-0.6758, 1.3259, 0.1472]^{\top}$. \ref{vis1}: The noisy sample consisting of examples with noisy labels $+1$/$-1$ (blue circles/red x-marks) and the classification boundaries. \ref{vis2}: Our training sample consisting of automatically-collected positive/negative examples (blue circles/red x-marks) and actively-labeled positive/negative examples (blue plus/red pentagrams). Observe that most of automatically-collected distilled examples are correctly collected. \ref{vis3}: A contour plot showing distilled examples' importance $\beta$, in which warmer color indicates greater importance. It demonstrates that actively-labeled examples are given greater importance than automatically-collected distilled examples, which is consistent with our analysis. \ref{vis4}: The test sample and classification boundaries.
}\label{vis}
\end{figure*}
Evaluations of our algorithm are conducted on both synthetic and real-world datasets. In our experiments, logistic regression is used for both training $\hat{f}$ and estimating $\tilde{\eta}$. For KMM, we always use the Gaussian kernel $k(\mathbf{x}_{i},\mathbf{x}_{j})=\exp(-\sigma\|\mathbf{x}_{i}-\mathbf{x}_{j}\|^{2})$ and the value of $\sigma$ is set as $\sigma=1$ for evaluations on synthetic datasets and $\sigma=0.01$ for evaluations on real-world datasets. The setup of parameters $\epsilon$ and $B$ is the same as that of \citet{huang2007correcting}, i.e., $\epsilon=(\sqrt{m}-1)/\sqrt{m}$ and $B=1000$. In this section, each entry in the tables is the result averaged over 1000 trials.
\subsection{The Case Where $\rho_{\text{+1max}}$ and $\rho_{\text{-1max}}$ Are Known}\label{sec72}
In this subsection, $\rho_{\text{+1max}}$ and $\rho_{\text{-1max}}$ are assumed to be known, and the evaluated methods are as follows:
\begin{enumerate}[label=\alph*)]
\setlength{\itemsep}{0pt}
\setlength{\parsep}{0pt}
\setlength{\parskip}{0pt}
\setlength{\topsep}{0pt}
\item ``clean'': Train the classifier with clean examples.
\item ``noisy'': Train the classifier with noisy examples.
\item ``auto'': Train the classifier with automatically-collected distilled examples.
\item ``auto+act'': Train the classifier with automatically-collected and manually-labeled distilled examples without importance reweighting. This can be viewed as a simplified version of our Algorithm \ref{algo2}.
\item ``Algo.\ 1'': Train the classifier by our Algorithm \ref{algo2}.
\item ``noisy+act'': Add the $n_\text{act}$ examples actively-labeled by ``auto+act'' into the noisy training sample and remove the corresponding noisy ones, and then train the classifier.
\end{enumerate}
Note that ``noisy+act'' is used as a baseline for comparisons with our ``auto+act''. The share of actively-labeled examples is to make comparisons fair.
\subsubsection{Evaluations on Synthetic Datasets}\label{sec721}

\begin{table*}[!tbp]
\footnotesize

\centering
\captionof{table}{Means and Standard Deviations (Percentage) of Classification Accuracies of Different Classifiers
}\label{tb3}
\begin{tabular}{ccccccccc}
\toprule
dataset & ($\rho_{\text{+1max}}$, $\rho_{\text{-1max}}$,$ n_\text{act}$ ) & clean & noisy & auto &  noisy+act & \tabincell{c}{auto+act \\ (ours)}  &\tabincell{c}{Algo.\ 1 \\ (ours)} \\
\midrule

\multirow{4}*{\tabincell{c}{Synthetic \\ Dataset}}&(0.25, 0.25, 3) & \multirow{4}*{99.73$\pm$0.17}&  98.55$\pm$1.28 & 99.13$\pm$0.83 & 98.56$\pm$1.27  & 99.28$\pm$0.70& \textbf{99.30$\pm$0.68} \\

~&(0, 0.49, 3) & ~& 90.06$\pm$9.43 & 97.42$\pm$2.75 &  90.06$\pm$9.43 & 97.75$\pm$2.38&   \textbf{98.01$\pm$2.10}  \\

~&(0.25, 0.49, 3)& ~&  92.59$\pm$8.64  & 97.95$\pm$2.70  &   92.60$\pm$8.63& 98.69$\pm$1.67& \textbf{98.90$\pm$1.37}   \\

~&(0.49, 0.49, 3) & ~&  { 88.57$\pm$10.74 }  & { 89.52$\pm$18.96}  & 88.74$\pm$10.60& 98.16$\pm$2.57& \textbf{98.43$\pm$2.29} \\

\midrule

\multirow{7}*{\tabincell{c}{UCI \\ Image}}&(0.1, 0.3, 20) & \multirow{7}*{83.10$\pm$1.36}&   81.16$\pm$2.21  &  81.39 $\pm$1.83  &   81.15$\pm$2.21   &  81.78$\pm$1.72&  \textbf{82.09$\pm$1.71}      \\

~&(0.3, 0.1, 20)& ~& 78.88$\pm$3.07  & 79.82$\pm$2.76  &     79.09$\pm$2.99     & 80.69$\pm$2.47&     \textbf{81.60$\pm$2.15}   \\

~&(0.2, 0.4, 20) & ~&  78.94$\pm$3.10  & 77.96$\pm$3.16  &    78.98$\pm$3.12    &  79.46$\pm$2.72&    \textbf{81.08$\pm$2.32} \\

~&(0.4, 0.2, 20) & ~&  75.80$\pm$4.08  & 76.30$\pm$3.67  &    76.14$\pm$3.96    &  78.44$\pm$3.37&  \textbf{80.35$\pm$2.70} \\

~&(0.3, 0.3, 20) & ~& 79.02$\pm$2.90  & 77.48$\pm$3.23 &    79.21$\pm$2.82    &    79.16$\pm$2.88&  \textbf{80.97$\pm$2.34}  \\

~&(0.4, 0.4, 20) & ~&  74.72$\pm$4.06 & 71.86$\pm$5.22 &   75.03$\pm$3.97    & 76.27$\pm$4.11&  \textbf{78.31$\pm$3.63}\\

~&(0.5, 0.5, 20) & ~&  68.72$\pm$5.91 & 64.49$\pm$7.39 &   69.19$\pm$5.80    &    73.64$\pm$4.95 & \textbf{75.64$\pm$4.69}\\

\midrule

\multirow{7}*{\tabincell{c}{USPS \\ (6vs8)}}&(0.1, 0.3, 20) & \multirow{7}*{98.07$\pm$0.52}& 89.00$\pm$1.84 & 93.55$\pm$1.47 &  89.01$\pm$1.82  & 93.72$\pm$1.43&   \textbf{93.74$\pm$1.44}   \\

~&(0.3, 0.1, 20)& ~& 89.15$\pm$1.78 & 93.64$\pm$1.50  &   89.30$\pm$1.77   & 93.82$\pm$1.41&   \textbf{93.83$\pm$1.44}    \\

~&(0.2, 0.4, 20) & ~&  86.40$\pm$2.31  & 91.45$\pm$1.91 &   86.46$\pm$2.31   &  91.65$\pm$1.85&  \textbf{ 91.67$\pm$1.86 }  \\

~ &(0.4, 0.2, 20) & ~&  86.45$\pm$2.27  & 91.51$\pm$1.92  &   86.67$\pm$2.20   &    91.74$\pm$1.86 & \textbf{91.77$\pm$1.88}    \\

~&(0.3, 0.3, 20) & ~& 87.01$\pm$2.13  & 91.74$\pm$1.83 &   87.15$\pm$2.08   & 91.97$\pm$1.73&  \textbf{91.98$\pm$1.77}  \\

~&(0.4, 0.4, 20) & ~&  82.84$\pm$2.81 & 87.77$\pm$2.74 &  83.08$\pm$2.77  & \textbf{88.36$\pm$2.55} & 88.31$\pm$2.60    \\

~&(0.5, 0.5, 20) & ~&  77.73$\pm$3.96 & 82.22$\pm$4.20 &   78.06$\pm$3.88    &      \textbf{83.35$\pm$3.90} & 83.19$\pm$3.92   \\
\bottomrule
\end{tabular}
\end{table*}

First, we perform evaluations on 2D  synthetic datasets that are linearly inseparable.

In each trial, positive examples and negative examples are sampled from two 2D normal distributions $\mathcal{N}_{1}(\mathbf{u},\mathbf{I})$ and $\mathcal{N}_{2}(-\mathbf{u},\mathbf{I})$ respectively, where $\mathbf{u}=[-2,2]^{\top}$ and $\mathbf{I}\in\mathbb{R}^{2\times2}$ is the identity matrix. Each 2D datapoint $\left(\mathbf{x}(1),\mathbf{x}(2)\right)$'s feature vector is $[1,\mathbf{x}(1),\mathbf{x}(2)]^{\top}$ where $1$ act as an intercept term.
We generate bounded instance- and label-dependent label noise at random via
\begin{align}\label{noise}
\left\{
\begin{aligned}
\rho_{+1}(\mathbf{x}) = \rho_{+1\text{max}}\cdot S\left(\mathbf{w}_{+1}^{\top}\mathbf{x}\right),\\
\rho_{-1}(\mathbf{x}) = \rho_{-1\text{max}}\cdot S\left(\mathbf{w}_{-1}^{\top}\mathbf{x}\right),\\
\end{aligned}
\right.
\end{align}
where elements of $\mathbf{w}_{+1},\mathbf{w}_{-1}$ are i.i.d.\ sampled from the standard normal distribution in each trial and $S(\cdot)$ denotes the sigmoid function $S(x)=\frac{1}{1+\exp{(-x)}}$.

The performances of evaluated methods under different ($\rho_{\text{{+1max}}}$, $\rho_{\text{{-1max}}}$, $n_\text{act}$) are shown in Table \ref{tb3}. Note that the standard deviations of some results are large, since in different trials label noise is generated by different noise rates functions. It is shown that our ``auto+act'' does not only always significantly outperform the baseline in terms of average classification accuracies, but also achieves smaller standard deviations compared to ``noisy'' and ``noisy+act''.

The procedure of Algorithm \ref{algo2} is visualized in Fig.\ \ref{vis}.
\begin{table*}[!t]
\centering
\footnotesize
\caption{Means and Standard Deviations (Percentage) of Classification Accuracies of Different Classifiers 
}
\label{tb5}
\begin{tabular}{cccccc}
\toprule
dataset &$\left(\rho_{\text{+1max}}, \rho_{\text{-1max}},n_\text{act}\right)$ & \tabincell{c}{auto \\ w/o \text{$\rho_{\text{$\pm$1max}}$}} & \tabincell{c}{noisy+act \\ w/o \text{$\rho_{\text{$\pm$1max}}$}}  &  \tabincell{c}{Algo.\ 1 \\ w/o \text{$\rho_{\text{$\pm$1max}}$}} \\
\midrule

\multirow{4}*{\tabincell{c}{Synthetic \\ Dataset}}&(0.25, 0.25, 3)  & \multirow{1}*{99.54$\pm$0.31} & 98.62$\pm$1.25  &  \textbf{99.61$\pm$0.33}  \\

~&(0, 0.49, 3) & \multirow{1}*{98.20$\pm$1.35} & 89.67$\pm$9.67 & \textbf{99.16$\pm$0.72} \\

~&(0.25, 0.49, 3) & \multirow{1}*{99.10$\pm$2.24}  &   92.54$\pm$9.00  & \textbf{99.41$\pm$0.74}  \\

~&(0.49, 0.49, 3) &  \multirow{1}*{92.36$\pm$19.09}  & 89.10$\pm$9.62 & \textbf{99.23$\pm$1.02}   \\
\midrule

\multirow{7}*{\tabincell{c}{UCI \\ Image}}&(0.1, 0.3, 20) &   78.65$\pm$2.64  &   81.19$\pm$2.16   &     \textbf{81.35$\pm$2.45}   \\

~&(0.3, 0.1, 20)& 75.00$\pm$3.84  & 79.25$\pm$3.06 &     \textbf{80.38$\pm$2.85}\\

~&(0.2, 0.4, 20) & 75.52$\pm$4.42  &    78.96$\pm$2.97    &    \textbf{79.51$\pm$3.18}  \\

~&(0.4, 0.2, 20) & 71.38$\pm$5.03  &    76.26$\pm$3.79    &     \textbf{78.63$\pm$3.56} \\

~&(0.3, 0.3, 20)  & 73.90$\pm$5.11 &    79.06$\pm$2.74   &     \textbf{79.30$\pm$3.30} \\

~&(0.4, 0.4, 20) & 69.20$\pm$5.64 &      75.01$\pm$3.78   &  \textbf{76.85$\pm$4.52}  \\

~&(0.5, 0.5, 20) &   64.61$\pm$6.87 &   69.45$\pm$5.91    &   \textbf{74.51$\pm$5.43}  \\
\midrule

\multirow{7}*{\tabincell{c}{USPS \\ (6vs8)}}&(0.1, 0.3, 20) & 94.96$\pm$1.24 &  89.03$\pm$1.82  &  \textbf{95.12$\pm$1.20}   \\

~&(0.3, 0.1, 20) & 95.02$\pm$1.28  &   89.34$\pm$1.79   &   \textbf{95.15$\pm$1.26}    \\

~&(0.2, 0.4, 20)  & 92.44$\pm$1.76  &   86.34$\pm$2.29   &  \textbf{92.73$\pm$1.69}  \\

~&(0.4, 0.2, 20  & 92.55$\pm$1.89  &   86.55$\pm$2.21   &   \textbf{92.83$\pm$1.78}  \\

~&(0.3, 0.3, 20) & 93.15$\pm$1.69 &87.03$\pm$1.95   &  \textbf{93.46$\pm$1.63}     \\

~&(0.4, 0.4, 20) &  88.87$\pm$2.71 &   83.00$\pm$2.78  &    \textbf{89.35$\pm$2.61}  \\

~&(0.5, 0.5, 20) &  82.52$\pm$4.06 &   77.95$\pm$3.86    & \textbf{83.40$\pm$3.87}  \\
\bottomrule
\end{tabular}
\end{table*}
\begin{figure*}[!t]
\centering
\subfigure[]{
  \begin{minipage}[t]{0.32\linewidth} 
    \centering
    \includegraphics[width=\linewidth]{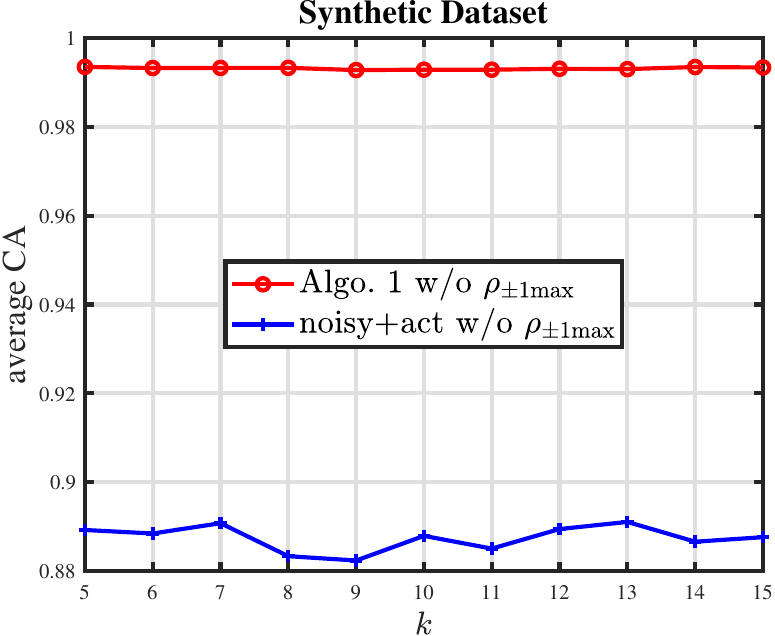}
    \label{fig71}
  \end{minipage}%
  }
  \subfigure[]{
  \begin{minipage}[t]{0.32\linewidth}
    \centering
    \includegraphics[width=\linewidth]{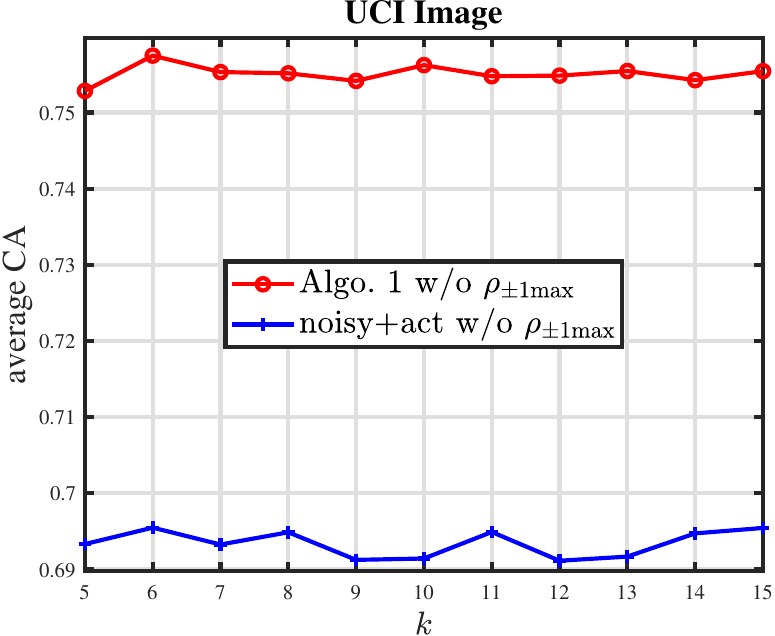}
    \label{fig72}
  \end{minipage}
  }
  \subfigure[]{
      \begin{minipage}[t]{0.32\linewidth} 
    \centering
    \includegraphics[width=\linewidth]{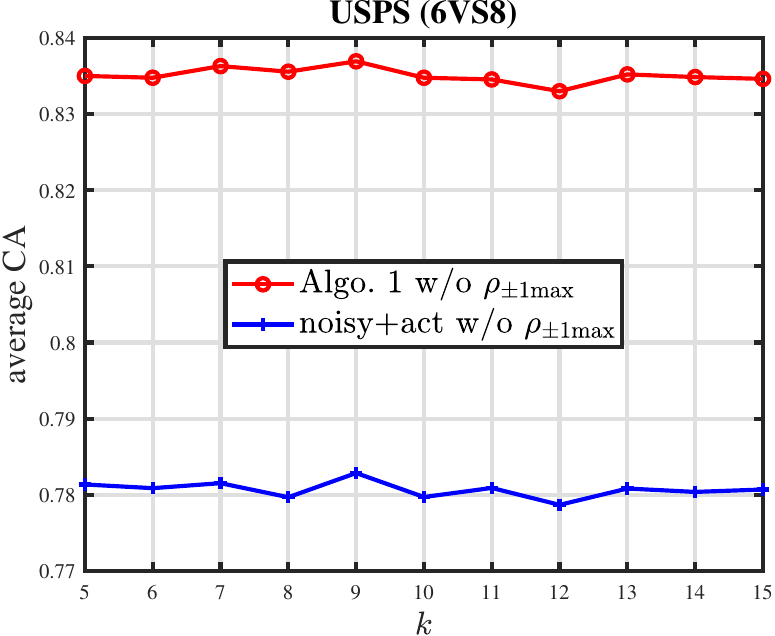}
 \label{fig73}
  \end{minipage}
  }
\caption{Curves illustrating the average classification accuracy (CA) of different classifiers learned without knowledge of $\rho_{+1\text{max}}$ and $\rho_{-1\text{max}}$ versus the hyperparameter $k$ on three datasets. The settings of ($\rho_{+1\text{max}}$, $\rho_{-1\text{max}}, n_\text{act}$) in \ref{fig71} and \ref{fig72}/\ref{fig73} are respectively (0.49, 0.49, 5) and (0.5, 0.5, 20). Each result in this figure is averaged over $1000$ trials.}\label{fig8}
\end{figure*}
\subsubsection{Evaluations on Real-World Datasets}\label{sec722}

Second, we conduct evaluations on two public real-world datasets: the image dataset from the UCI repository provided by Gunnar R{\"a}tsch\footnote{\url{http://theoval.cmp.uea.ac.uk/matlab}} and the USPS handwritten digits dataset\footnote{\url{http://www.cs.nyu.edu/~roweis/data.html}} \citep{hull1994database}. The UCI Image dataset is composed of 1188 positive examples and 898 negative examples. As for the USPS dataset, We use images of ``6'' and ``8''  as positive and negative examples respectively, and each class has 1100 examples. In each trial, all feature vectors are standardized so that each element has roughly zero mean and unit variance, and examples are randomly split, $75\%$ for training and $25\%$ for testing. Label noise is generated in the same way with evaluations on synthetic datasets by Eq.\ (\ref{noise}).

Evaluation results are shown in Table \ref{tb3}. It is shown that our algorithm still outperforms the baselines.

\subsection{The Case Where $\rho_{\text{+1max}}$ and $\rho_{\text{-1max}}$ Are Unknown}\label{sec73}
In this section, we evaluate variants of ``auto'', ``noisy+act'' and ``Algo.\ 1'', which employ the approach proposed in Sec.\ \ref{sub553} to collect distilled example without knowledge of $\rho_{\pm1\text{max}}$, on both synthetic and real-world datasets. The evaluated methods are denoted as
``auto w/o $\rho_{\pm1\text{max}}$'', ``noisy+act w/o $\rho_{\pm1\text{max}}$'' and ``Algo.\ 1 w/o $\rho_{\pm1\text{max}}$''.

The setup of experiments is same with setup in Sec.\ \ref{sec72}. The only newly introduced hyperparameter is $k$. We set $k=10$ for all evaluations on three datasets, and avoid dataset specific tuning.

Results are listed in Table \ref{tb5} and show that our ``Algo.\ 1 w/o $\rho_{\pm1\text{max}}$'' outperforms the baseline. Comparing results in Table \ref{tb5} and results in Sec.\ \ref{sec72}, {we are surprised to observe that the performance of ``Algo.\ 1 w/o $\rho_{\pm1\text{max}}$'' is comparable with and sometimes even better than that of ``Algo.\ 1''}.

To investigate the sensitivity of our algorithm to $k$, we plot the performance curves w.r.t.\ the variation of $k$  on three datasets in Fig. \ref{fig8}. It is shown that our algorithm is robust to the selection of $k$ on different datasets.

\section{Conclusion}\label{sec7}
In this paper, we focus on learning with BILN, which is a more general case of label noise than those have been well-studied. We propose a learning algorithm and theoretically establish its statistical consistency and a performance bound. Empirical evaluations on synthetic and real-world datasets show effectiveness of the proposed algorithm. 

In future, we will explore the combination of our algorithm and more complicated models (e.g., deep neural networks) for real-world label noise tasks.

\section*{Acknowledgements}
This work was supported in part by Australian Research Council Projects, i.e., DE-190101473, FL-170100117, DP-180103424, IH-180100002, IC-190100031, LE-200100049. We thank the anonymous reviewers for their constructive comments.

\bibliography{ref}
\bibliographystyle{icml2020}


\end{document}